# Face Verification System based on Integral Normalized Gradient Image (INGI)

V. Karthikeyan
Assistant Professor,
Department of ECE,
SVS College of Engg,
Coimbatore, India

M. Divya
U.G Student,
Department of ECE,
SVS College of Engg,
Coimbatore, India

C.K. Chithra
U.G Student,
Department of ECE,
SVS College of Engg,
Coimbatore, India

K. Manju Priya
U.G Student,
Department of ECE,
SVS College of Engg,
Coimbatore, India

## ABSTRACT
Face Recognition is to refine the notion of a biometric imposter, and show that the traditional measures of identification and verification performance. Recognition algorithm performs scores on disjoint populations to institute a means of computing and display distribution-free estimates of the dissimilarity in verification vs. false alarm performance. The proposed face recognition system consists of an Illumination Insensitive Preprocessing Method, A Hybrid Fourier-Based Facial Feature Extraction, and Score Fusion Scheme. In pre-processing stage, a figure is normalized and integrated called "integral normalized gradient image". Then, in feature extraction of complementary classifiers, for multiple face models hybrid Fourier features is applied. Multiple face models are generated by normalized face images that have different eye distances. Finally, to combine scores from multiple complementary classifiers, a log likelihood ratio-based score fusion scheme is used. The goal of the Face Recognition Grand Challenge (FRGC) is to enhance the performance of face recognition algorithms by the order of magnitude.

## 1. INTRODUCTION
In the past decades, many appearance-based methods have been proposed to handle this problem, and new theoretical insights as well as good recognition results have been reported. In the proposed the verification of the face in different climatic conditions, this paper focuses mainly on the issue of robustness to lighting variations The major issue for face recognition is to ensure recognition accuracy for a large data set captured in various conditions. The FRGC is intended to achieve this performance goal along with a data corpus of images. The data consists of 3D scans and high resolution still images which are taken under controlled and uncontrolled conditions. The FRGC data set contains face images collected in different settings with two different facial expressions (neutral versus smiling) taken for several months. A subject session is the set of all images of a person taken each time a person's biometric data is collected [5]. The FRGC data for a subject session consists of controlled still images, uncontrolled still images, and one three-dimensional image. The controlled images were taken in a studio setting, full frontal facial images taken under two lighting conditions and with facial expressions [5]. The uncontrolled images were taken in varying illumination conditions such as outdoors and hallways. The uncontrolled environmental problems can be overcome by systematic approach that combines multiple classifiers with complementary features instead of improving the accuracy of a single classifier [4].

## 2. PRE-PROCESSING
Normalization based approaches seek to reduce the image to a more "canonical" form in which the illumination variations are suppressed. Histogram equalization is one simple example of this method. This is a post-processing transform of similarity scores that may utilize the fact that the gallery, unlike the target set, contains only one image per person by definition. Normalization is defined as a function, f: $R_N \to R_{N'}$ mapping s to a new vector, t. for an algorithm that uses normalization, the final performance scores are computed over these transformed values [9]. The probe and gallery images must be the same size and is normalized to line up the eyes and mouth of the subjects within the image. The approach is then used to reduce the dimension of the data by means of data compression basics and reveals the most effective low dimensional structure of facial patterns. This reduction in dimensions removes information that is not useful and precisely decomposes the face structure into orthogonal components known as Eigen faces [4]. The changes in terms of different illuminations among the same person are greater than those of different persons among the same illumination [3].

### 2.1 Integral Normalized Gradient Image
The pre-processing SQI (self-quotient image) method could remove most of the shaded parts of a face image but results in some noise and halo effects in the step regions [2]. In this paper, the illumination-insensitive image integral normalized gradient image (INGI) method is proposed to trounce the unexpected illumination changes in face recognition with restricted side effects such as image noise and the halo effect. Based upon intrinsic and extrinsic factor definitions, we first normalize the gradients with a smoothed image and then integrate the division results with the anisotropic diffusion method. Assumptions made as 1) most of the intrinsic factor is in the high spatial frequency domain, and 2) most of the extrinsic factor is in the low spatial frequency domain. Considering the first assumption, one might use a high-pass filter to extract the intrinsic factor, but it has been proved that this kind of filter is not robust to illumination variations. In addition, a high-pass filter may remove some of the useful intrinsic factor. Hence, we propose an alternative approach, namely, employing a gradient operation. The gradient operation is written

$$\begin{aligned}\nabla_\chi &= \nabla(\rho \sum_i n^T . s_i) \\ &= (\nabla_\rho) \sum_i n^T . s_i + \rho \nabla(\sum_i n^T . s_i) \\ &= (\nabla_\rho) \sum_i n^T . s_i \\ &= (\nabla_\rho) W \end{aligned} \quad (1)$$

To overcome the illumination sensitivity, we normalized the gradient map with the following equation:

$$N = \frac{\nabla_\chi}{W} = \frac{(\nabla_\rho)W}{W} = \nabla_\rho \quad (2)$$

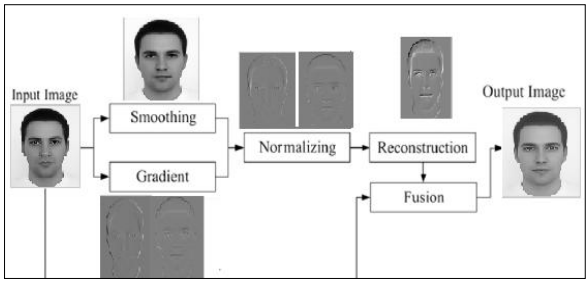

**Fig 1: INGI based pre-processing**

After the normalization, the texture information is transferred to normalized image $N=\{N_x, N_y\}$ is still not apparent enough.

To recover the rich texture and remove the noise at the same time, we integrate the normalized gradients $N_x$ and $N_y$ with the anisotropic diffusion method which we explain in the following, and finally acquire the reconstructed image $\chi_r$ [4].

## 2.2 Reconstruction of image

The INGI procedure is to improve a grayscale image from normalized gradient maps. If a grayscale value of one point in an image, we can estimate the grayscale of any point by an integration method, such as an iterative isotropic diffusion method. Our intent is to attain the basis images by using each person's pictures respectively; using the entire training descriptions of the database. We obtained the features which would be employed to reconstruct the images by mapping the test images to the basis images. Adopting the minimum reconstruction error is the major issue in the paper.

$$\chi_{r(i,j)}^t = \frac{1}{4}[(\chi_{r(i,j)}^{t-1} + N_{N,(i,j)}) + (\chi_{r(i,j)}^{t-1} + N_{S,(i,j)})N_{W,(i,j)}) + (\chi_{r(i,j)}^{t-1} + N_{E,(i,j)})] \quad (3)$$

Where **t** is an iteration number $\chi_r^t$, is the iterative reconstructed image and usually
$\chi_r^0 = 0$ [4].

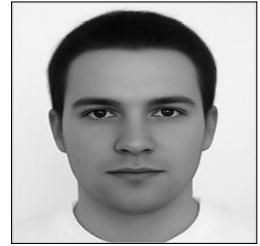 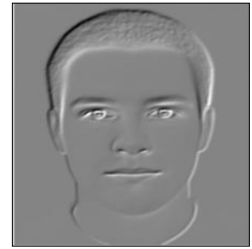

(a)            (b)

**Fig 2 (a): Original image     Fig 2 (b): reconstructed image**

The reconstruction is realized to project face images into a low-dimensional vector space and to reconstruct the respective reference images from the projected vectors [2].

## 3. FEATURE EXTRACTION
### 3.1 Kernel PCA

Kernel Principal Component Analysis (KPCA) is used for feature extraction from hyper spectral remote sensing data. Accuracy is improved when compared to original approach which used conventional principal component analysis [8]. KPCA encodes the pattern information based on second order dependencies, i.e., pixel wise covariance among the pixels, and are insensitive to the dependencies of multiple (more than two) pixels in the patterns. Since the eigenvectors in PCA are the orthonormal bases, the principal components are uncorrelated. In other words, the coefficients for one of the axes cannot be linearly represented from the coefficients of the other axes.     In this paper, an alternative approach to data fusion is discussed. The PCA is optimal for representation under some simple

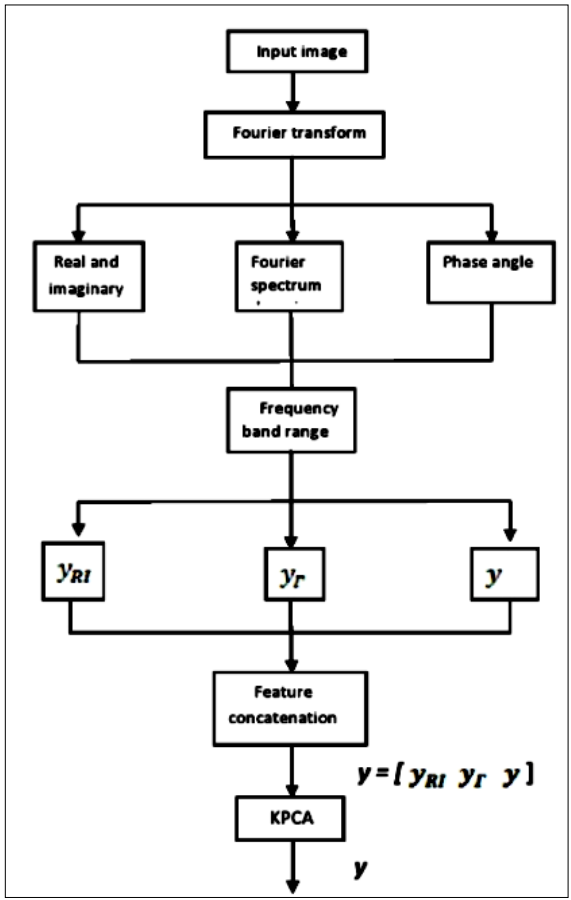

**Fig 4: Feature extraction by KPCA**

The *n* observed variable, *i.e.*, the spectral channels **x ε R***n*, result from a linear transformation of m latent variables Gaussian distributed and thus it is possible to recover the latent variable,[8] *i.e.*, the principal components, from the observed one by solving the following Eigen value problem:

$$\lambda v = \Sigma_x v \text{ subject to } \|v\|_2 = 1 \qquad (4)$$

The PCA only relies on second order statistics and theoretical limitations for hyper spectral data analysis. Since the PCA does not handle all the spectral information, another unsupervised feature extraction is proposed, namely the Kernel PCA (KPCA) .

### 3.2 Hybrid Fourier Feature

More information is extracted by KPCA and is more robust to non-Gaussian noise, since a reasonable number of features are extracted. Matching algorithm is a overview of string algorithm. The transformation of a 2Dpattern into a 1Dpattern usually results in a failure of information [15]. Attributed graph matching has often been discussed and advocated in the context of object recognition. Our pattern detection system is based on a unusual form of attributed graph matching which resembles elastic matching, an attributed graph in the model domain, encoding an object, is locally indistinct to handle with deformations and change in perception. In the mathematical literature, the relation to be established in graph matching is precise isomorphism. This is too restrictive for the purposes of object recognition [12]-[14].We have three different Fourier feature domains, namely, the real and imaginary component (**RI**) domain, Fourier spectrum domain($\Gamma$), and phase angle domain($\Phi$). We present how we apply the three frequency band selections $B_1, B_2$ and $B_3$ to the three Fourier feature domains [4].

### 3.3 Fourier Transform Based Spectrum

The Fourier transform has played a key role in image processing applications. The feature is extracted using Fourier domain; we apply the Fourier transform to a spatial face image.

$$F_{(u,v)} = R_{(u,v)} + jI_{(u,v)} \qquad (5)$$

Where $R_{(u,v)}$ and $I_{(u,v)}$ are the real and imaginary components of $F_{(u,v)}$, respectively

To extract an image, the magnitude coefficients are widely used instead of the phase values. This is largely because a little spatial displacement in an image will change the phase values drastically while the magnitude varies smoothly when there is no compensator for the phase shift [4]. To extract an image, the magnitude coefficients are widely used instead of the phase values. This is largely because a little spatial displacement in an image will change the phase values drastically while the magnitude varies smoothly when there is no compensator for the phase shift [4]

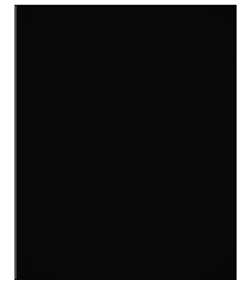

**Fig 3: Illustration of feature extraction extracted image**

The magnitude function called the Fourier spectrum is

$$|F_{(u,v)}| = [R^2_{(u,v)} + I^2_{(u,v)}]^{1/2} \qquad (6)$$

The phase angle is

$$\Phi_{(u,v)} = \tan^{-1}\left[\frac{I_{(u,v)}}{R_{(u,v)}}\right] \qquad (7)$$

To avoid the angular problem in phase angle, $2\pi \approx 0$, we use the cosine values

$$\Phi_{(u,v)} = \cos\left(\tan^{-1}\left[\frac{I_{(u,v)}}{R_{(u,v)}}\right]\right) \qquad (8)$$

KPCA does not require solving a nonlinear optimization problem which is expensive computationally and the validity of the solution as optimal is typically a concern. KPCA only requires the solution of an eigen value problem. This reduces to using linear algebra to perform PCA in an arbitrarily large, possibly infinite dimensional, feature space. The kernel "trick" greatly simplifies calculations in this case. An additional advantage of KPCA is that the number of components does not have to be specified in advance. KPCA is a useful generalization that can be applied to these domains where nonlinear features require a nonlinear feature extraction tool. Nowadays KPCA algorithm is been used for real earth science data such as the sea surface temperature (SST) or normalized difference vegetation index (NDVI). The resulting information from KPCA can be correlated with signals such as the Southern Oscillation Index (SOI) for determining relationships with the El Nino phenomenon. KPCA can be used to discover nonlinear correlations in data that may otherwise not be found using standard PCA[8]. The information generated about a data set using KPCA captures nonlinear features of the data. These features correlated with known spatial-temporal signals can discover nonlinear relationships. KPCA offers improved analysis of datasets that have nonlinear structure. To better understand the link and the difference between PCA and KPCA, one must note that the eigenvectors of $\Sigma_x$ can be obtained from those of $XX^T$, where $X = [x_1, x_2, \ldots, x_l]^T$ Consider the Eigen value problem:

$$U = XX^T u, \text{ subject to } \|u\|_2 = 1. \tag{9}$$

The left part is multiplied by $X^T$ giving

$$X^T u = X^T X X^T u,$$
$$X^T u = (l-1) \Sigma_x X^T u, \tag{10}$$
$$X^T u = \Sigma_x X^T u,$$

Which is the Eigen value problem (1): $v = X^T u$. But $\|v\|^2 = u^T = XX^T u, = {}_\gamma u^T u = {}_\gamma \neq 1$.

Therefore, the eigenvectors of $\Sigma_x$ can be computed from Eigen vectors of $XX^T$ as $v = {}^{-0.5} X^T u$. The matrix $XX^T$ is equal to:

$$\begin{pmatrix} (x^1 x^1) & \cdots & (x^1 x^l) \\ \vdots & \ddots & \vdots \\ (x^l x^1) & \cdots & (x^l x^l) \end{pmatrix} \tag{11}$$

**Which is kernel matrix with a linear kernel: $k(x^i, x^j) = (x^i, x^j) R_n$. Using the kernel trick, K can be rewritten in a:**

$$\begin{pmatrix} <\Box(x^1), \Box(x^1)>H & \cdots & <\Box(x^1), \Box(x^l)>H \\ \vdots & \ddots & \vdots \\ <\Box(x^l), \Box(x^1)>H & \cdots & <\Box(x^l), \Box(x^l)>H \end{pmatrix} \tag{12}$$

From the above equation, the advantage of using KPCA is an appropriate projection of $R^n$ onto $\mathcal{H}$. The data should better match the PCA model. It is clear that the KPCA shares the same properties as the PCA, but in different space [8].

## 4. CONCLUSION

The face recognition algorithm is used to identify the face. The variation in performance is compared by two sets of data. A category of test images are saved in gallery, to compare with the train data. The train data is read and the size is validated. The test data is pre processed in order to remove unwanted effects. In score fusion technique Euclidian distance is calculated. The score value of test data is calculated and the value of training data is also calculated. The above two score values are compared and if the score matches above 85% then the test and training image are the same. The time taken for performing this performance check is also updated. A match score is a similarity score between two images of the same face. The primary goal of the paper is to encourage the development of face recognition algorithms that have performance an order of magnitude better than observed in previous results.

## 5. EXPERIMENTAL RESULTS

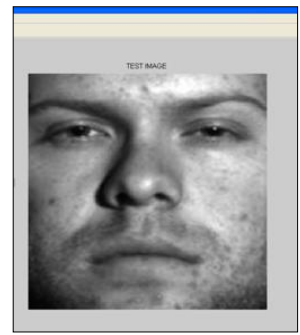

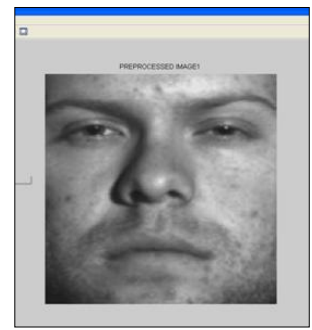

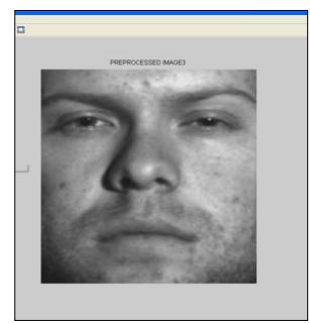

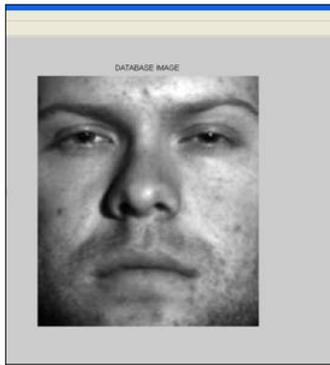